\documentclass[review]{elsarticle}
\usepackage{graphicx}
\usepackage{amsfonts}
\usepackage{amsmath}
\usepackage{multirow}
\usepackage{times}
\usepackage{epsfig}
\usepackage{amssymb}
\usepackage{mathtools}
\usepackage[table,xcdraw, dvipsnames]{xcolor}
\usepackage{xcolor}
\usepackage[caption=false]{subfig}

\journal{}

\begin{document}

\begin{frontmatter}

\title{PNL: Efficient Long-Range Dependencies Extraction with Pyramid Non-Local Module for Action Recognition}

\author[1]{Yuecong Xu\corref{cor1}%
\fnref{fn1}}
\ead{xuyu0014@e.ntu.edu.sg}
\author[1]{Haozhi Cao\fnref{fn1}}
\ead{haozhi001@e.ntu.edu.sg}
\author[1]{Jianfei Yang\fnref{fn1}}
\ead{yang0478@e.ntu.edu.sg}
\author[1]{Kezhi Mao\fnref{fn1}}
\ead{ekzmao@ntu.edu.sg}
\author[2]{Jianxiong Yin\fnref{fn2}}
\ead{jianxiongy@nvidia.com}
\author[2]{Simon See\fnref{fn2}}
\ead{ssee@nvidia.com}
\cortext[cor1]{Corresponding author}
\fntext[fn1]{School of Electrical and Electronic Engineering, Nanyang Technological University.}
\fntext[fn2]{NVIDIA AI Tech Centre.}
\address[1]{50 Nanyang Avenue, 639798, Singapore}
\address[2]{3 International Business Park Rd, \#01-20A Nordic European Centre, 609927, Singapore}

\begin{abstract}
Long-range spatiotemporal dependencies capturing plays an essential role in improving video features for action recognition. The non-local block inspired by the non-local means is designed to address this challenge and have shown excellent performance. However, the non-local block brings significant increase in computation cost to the original network. It also lacks the ability to model regional correlation in videos. To address the above limitations, we propose Pyramid Non-Local (PNL) module, which extends the non-local block by incorporating regional correlation at multiple scales through a pyramid structured module. This extension upscales the effectiveness of non-local operation by attending to the interaction between different regions. Empirical results prove the effectiveness and efficiency of our PNL module, which achieves state-of-the-art performance of $83.09\%$ on the Mini-Kinetics dataset, with decreased computation cost compared to the non-local block.
\end{abstract}


\end{frontmatter}


\section{Introduction}
\label{section:intro}

Action recognition has received considerable attention from the vision community in recent years \cite{herath2017going,yang2019asymmetric,carmona2018human,wang2018learning} thanks to its increasing applications in various fields, such as surveillance \cite{danafar2007action,xiang2008activity,li2017accurate} and smart homes \cite{wu2010multiview,ortis2017organizing,yang2018device} etc. Capturing long-range spatiotemporal dependencies have proven to play a key role in extracting effective video features for action recognition. Previously, feature extraction techniques such as SIFT \cite{lowe1999object}, GLOH \cite{mikolajczyk2005performance} and Dense Trajectory \cite{wang2011action} models such dependencies through hand-crafted features. More recently, convolutional and recurrent modules have replaced these hand-crafted features as the predominant modules for video feature extraction. However, each convolution or recurrent module extract spatiotemporal dependencies within spatial or temporal local regions. Therefore, it requires a stack of convolution or recurrent modules to model long-range spatiotemporal dependencies. Such strategy is computationally inefficient, while also causing difficulties in network optimization.

Inspired by the non-local means for image filtering task \cite{buades2005non,li2016novel}, the non-local neural network \cite{wang2018non} is proposed to address the challenge of capturing long-range dependencies directly. The proposed network captures long-range dependencies through direct modeling the correlation of each single pixel at any spatiotemporal location in a single module: non-local block. Without bells and whistles, the insertion of non-local block improves action recognition accuracy of existing networks, which proves the effectiveness of non-local block in extracting long-range dependencies. 

Despite the great improvement brought by the non-local block, the original non-local block has its own limitations. The original non-local block significantly increases the parameter size and computation cost of the network, thanks to the fact that the long-range dependencies is captured through pixel correlation. The increase in action recognition accuracy is at the cost of a significant decrease in computation efficiency of the network. 

\begin{figure}[!t]
    \centering
    \includegraphics[width=\linewidth]{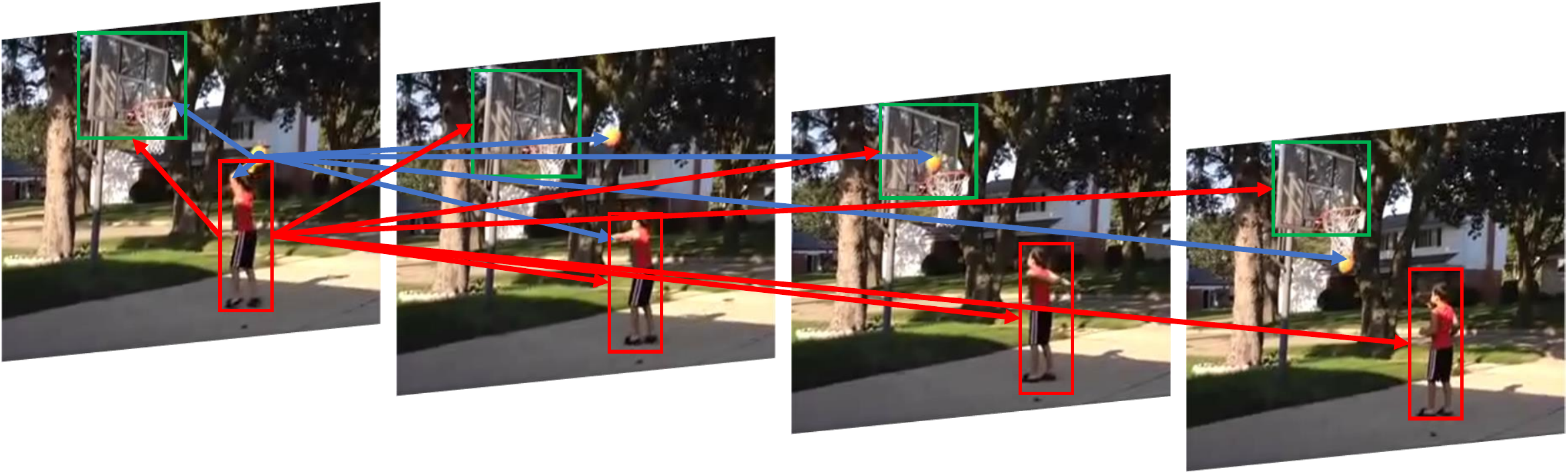}
    \caption{Illustration of utilizing regional correlation for action recognition. The original non-local block captures long-range spatiotemporal dependencies through pixel correlation, shown as blue arrows. The action ``playing basketball" could alternatively be recognized through regional correlation between the boy and the backboard, shown as red arrows.}
    \label{figure:intro:motivation}
\end{figure}

On the other hand, when we recognize action, it is more intuitive to focus on not only the correlation between each single pixel, but also on the correlation between larger regions of each frame, as can be shown in Figure~\ref{figure:intro:motivation}. To classify the action, we relate the boy with the backboard, which suggest a high possibility of the ``playing basketball" action. This is more efficient and intuitive than extracting pixel correlation that relates the basketball across frames, as well as pixel correlation that relates the basketball with the hands and elbows. 

To this end, to improve both the effectiveness and efficiency of the non-local block, we propose a novel long-range spatiotemporal dependencies extraction module: the Pyramid Non-Local (PNL) module. The proposed PNL module extends the original non-local block, and incorporates regional feature correlation at multiple scales through a pyramid structured module. The multi-scaled correlation are combined with a self-attentive combination function. Our main contributions are summarized as follows:
\renewcommand{\labelitemi}{$\ast$}
\begin{itemize}
    \item We propose a novel long-range spatiotemporal dependencies extraction module, Pyramid Non-Local (PNL) module. The PNL module extends the original non-local block through incorporating regional feature correlation at multiple scales. This extension upscales the effectiveness of non-local operation by attending to the interaction between different regions.
    \item We conduct comprehensive analysis over the computation cost required by our proposed PNL module. We further demonstrate its efficiency through comparing the computation cost of the PNL module against the original non-local block.
    \item We conduct extensive experiments on two action recognition benchmark datasets: Mini-Kinetics \cite{xie2018rethinking} and UCF101 \cite{soomro2012ucf101} with multiple frameworks utilizing our proposed PNL module. The results demonstrate that our proposed PNL module brings noticeable improvements over baseline methods and methods utilizing the original non-local block, while requiring less computation cost. Our network achieves state-of-the-art performance for the Mini-Kinetics dataset.
\end{itemize}

The rest of this paper is organized as follows. Related works for long-range spatiotemporal dependencies extraction in videos as well as pyramid structured neural networks are discussed in Section~\ref{section:related}. Subsequently, in Section~\ref{section:method}, we introduce and analyze the proposed Pyramid Non-Local module (PNL) in detail. After that, we present and analyze the experimental results of our proposed PNL module, with thorough ablation experiments on the design of PNL module and visualization of feature outputs. Finally, we conclude the paper and propose some future works in Section~\ref{section:conclusion}.

\section{Related Work}
\label{section:related}

\subsection{Capturing Long-range Spatiotemporal Dependencies}
\label{related:dependencies}
Capturing long-range spatiotemporal dependencies plays an important role in extracting effective video features. Previously, such dependencies are captured through hand-crafted features, extracted through algorithms such as SIFT \cite{lowe1999object}, GLOH \cite{mikolajczyk2005performance} and Dense Trajectory \cite{wang2011action}. The extracted hand-crafted features are effective, yet the extraction process is known to be computationally expensive and memory intensive. In addition, as extracting the hand-crafted features require pre-computation, the use of these algorithms prohibits fully end-to-end training of the network.

Convolutional and recurrent modules have become the predominant modules for video feature extraction with its good performance in action recognition task \cite{tran2015learning,richard2017bag}. However, vanilla convolutional and recurrent modules are both unable to capture long-range dependencies. Convolutional module captures spatiotemporal dependencies within spatial or temporal local regions. Whereas vanilla recurrent module, though designed for sequential data modeling, suffers from vanishing gradient problem which prevents it from capturing long-range dependencies \cite{pascanu2013difficulty}. Subsequently, various modules have been proposed in an attempt to better capture long-range dependencies. One notable module is the LSTM module \cite{hochreiter1997long}. It includes a 'memory cell' that can maintain information in memory for long periods of time such that long-range dependencies can be captured. LSTM have been utilized in various works for effective action recognition \cite{veeriah2015differential,sun2017lattice,shi2017learning}. However, LSTM module suffer from its large memory requirement and slow training speed while also prone to overfitting. Therefore current state-of-the-art action recognition models do not adopt LSTM module for long-range spatiotemporal dependencies.

More recently, inspired by the non-local means for image filtering task \cite{buades2005non,li2016novel}, the non-local block \cite{wang2018non} is introduced with the non-local neural network for capturing long-range dependencies. Subsequently, multiple variants of the non-local block have also been introduced. One of which is the compact generalized non-local operation \cite{yue2018compact}, which exploits cross-channel correlation on top of the original non-local operation. Another is the double-attention module \cite{chen20182} which computes correlation of features from a compact bag. Though both variants improves from the original non-local block, they have not considered the use of regional correlation at multiple scaled, which differs our proposed module with theirs.

\subsection{Pyramid Structured Neural Networks}
\label{related:pyramid}
Pyramid structured networks have proven to be effective in utilizing multi-scaled features for various tasks, including object detection \cite{lin2017feature}, pose estimation \cite{chen2018cascaded} and image segmentation \cite{seferbekov2018feature}. In the field of action recognition, pyramid structured networks have also been utilized to fuse spatial and temporal features \cite{wang2017spatiotemporal}. There are also works that utilize video frame inputs sampled at multiple temporal scales, such as the SlowFast network \cite{feichtenhofer2019slowfast}, which could be viewed as a pyramid structured network along the temporal dimension. In our work, the pyramid structure is utilized to extract multi-scaled regional correlation for capturing long-range spatiotemporal dependencies.

\section{Methodology}
\label{section:method}

The primary goal of our work is to develop a more effective and efficient module to extract long-range spatiotemporal dependencies. To achieve this, we propose to extend the non-local block \cite{wang2018non} through incorporating regional correlation. In this section, we introduce our proposed Pyramid Non-Local (PNL) module with detailed illustration of how it is extended from the original non-local block. We then provide solid proof over its higher computation efficiency compared to the original non-local block.


\begin{figure}[!t]
    \centering
    \subfloat[]{
        \includegraphics[width=\linewidth]{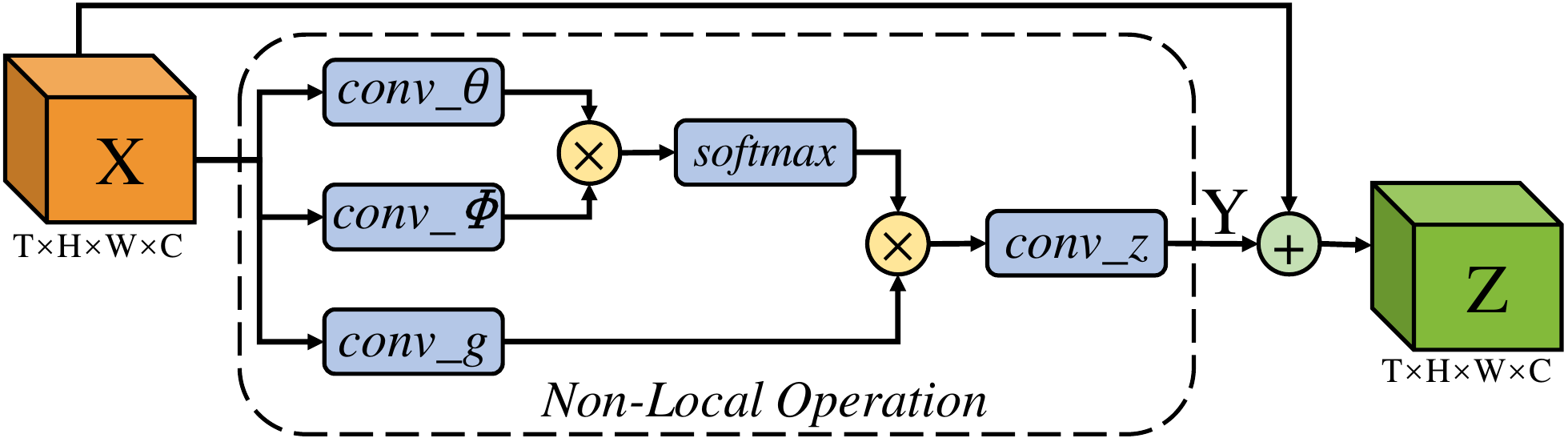}
        \label{figure:method:nl-pnl-a}
        }
    \small\smallskip\\
    \subfloat[]{
        \includegraphics[width=\linewidth]{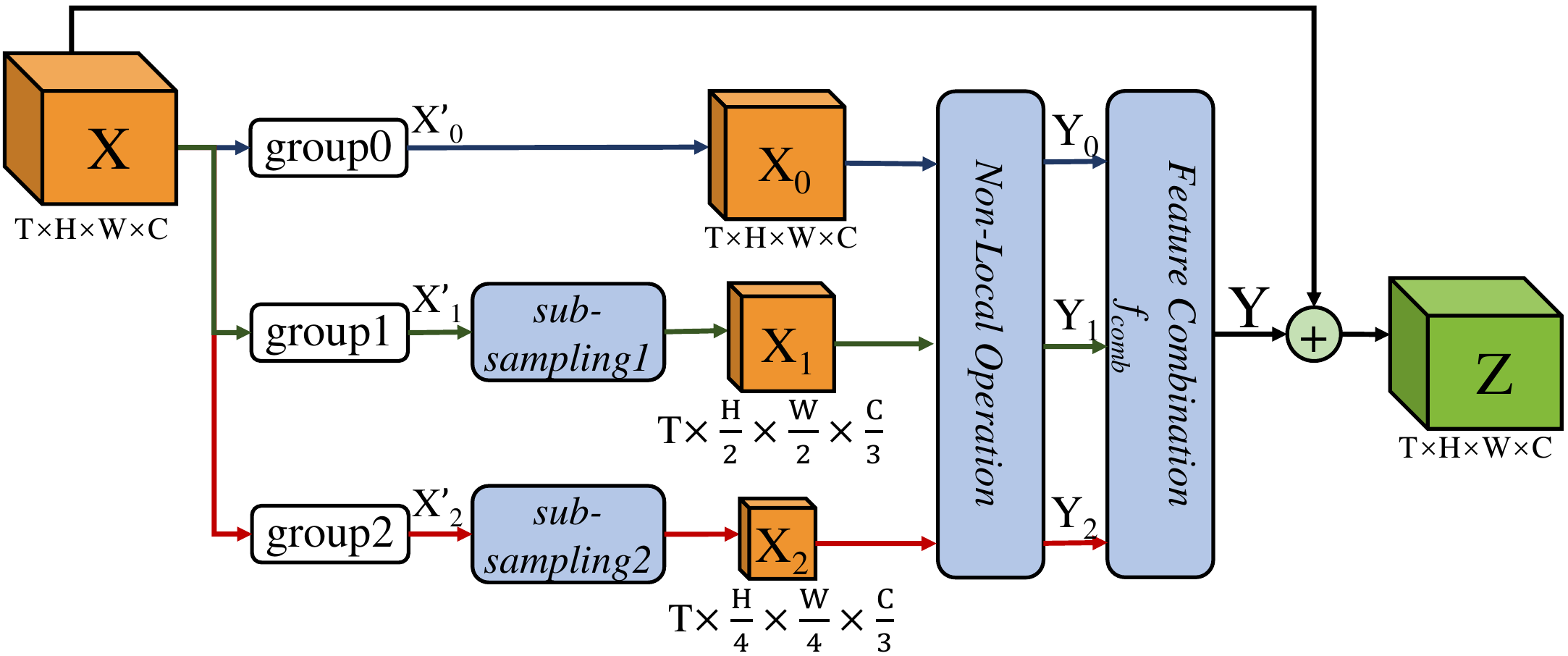}
        \label{figure:method:nl-pnl-b}
        }
    \caption{Comparison of the original non-local block (a) with our proposed PNL module (b). We present the case where the embedded Gaussian function is utilized for the non-local operation. The dimension of the input and output features are also presented, with the ``batch" dimension ignored.}
    \label{figure:method:nl-pnl}
\end{figure}

\subsection{Review of Non-Local Block}
\label{method:review-nl}
As our proposed module is built by extending the original non-local block, we begin by briefly reviewing the original non-local block as introduced in \cite{wang2018non}. The structure of the original non-local block is as shown in Figure~\ref{figure:method:nl-pnl-a}. Let the video input be denoted as $\mathbf{X}\in\mathbb{R}^{T\times H\times W\times C}$. where $T$, $H$, $W$ and $C$ denote the temporal length, height, width and number of channels of the video respectively. The original non-local block captures long-range spatiotemporal dependencies through non-local operation, which is a weighted sum of the correlation features at all positions, formulated as:
\begin{equation}
\label{eqn:method:orignal-nl}
\mathbf{y}_i = \frac{1}{\mathcal{C(\mathbf{X})}}\sum\limits_{\forall j}f(\theta(\mathbf{x}_i), \phi(\mathbf{x}_j))g(\mathbf{x}_j),
\end{equation}
where $\mathbf{y}_i$ is the output response $\mathbf{Y}$ at position i, while $\mathbf{x}_i$ and $\mathbf{x}_j$ are the input features at positions i and j. $\theta(\cdot)$, $\phi(\cdot)$ and $g(\cdot)$ are learnable transformations of the input features and are implemented as convolution layers with kernel size of $1\times1\times1$. Due to the convolutional implementation, we specify the transformations $\theta(\cdot)$, $\phi(\cdot)$ and $g(\cdot)$ to be $conv$\textunderscore$ \theta$, $conv$\textunderscore$ \phi$ and $conv$\textunderscore $g$ respectively. The pairwise function $f(\cdot,\cdot)$ computes the affinity between the input features at all space-time positions. The choice of the pairwise function $f(\cdot,\cdot)$ varies. Here we show the case where the embedded Gaussian version of $f(\cdot,\cdot)$ is adopted in Figure~\ref{figure:method:nl-pnl-a}. The final output of the non-local block $\mathbf{Z}$ is computed by adding the long-range dependencies $\mathbf{Y}$ from the non-local operation with the original input $\mathbf{X}$.

\begin{figure}[!t]
    \centering
    \includegraphics[width=\linewidth]{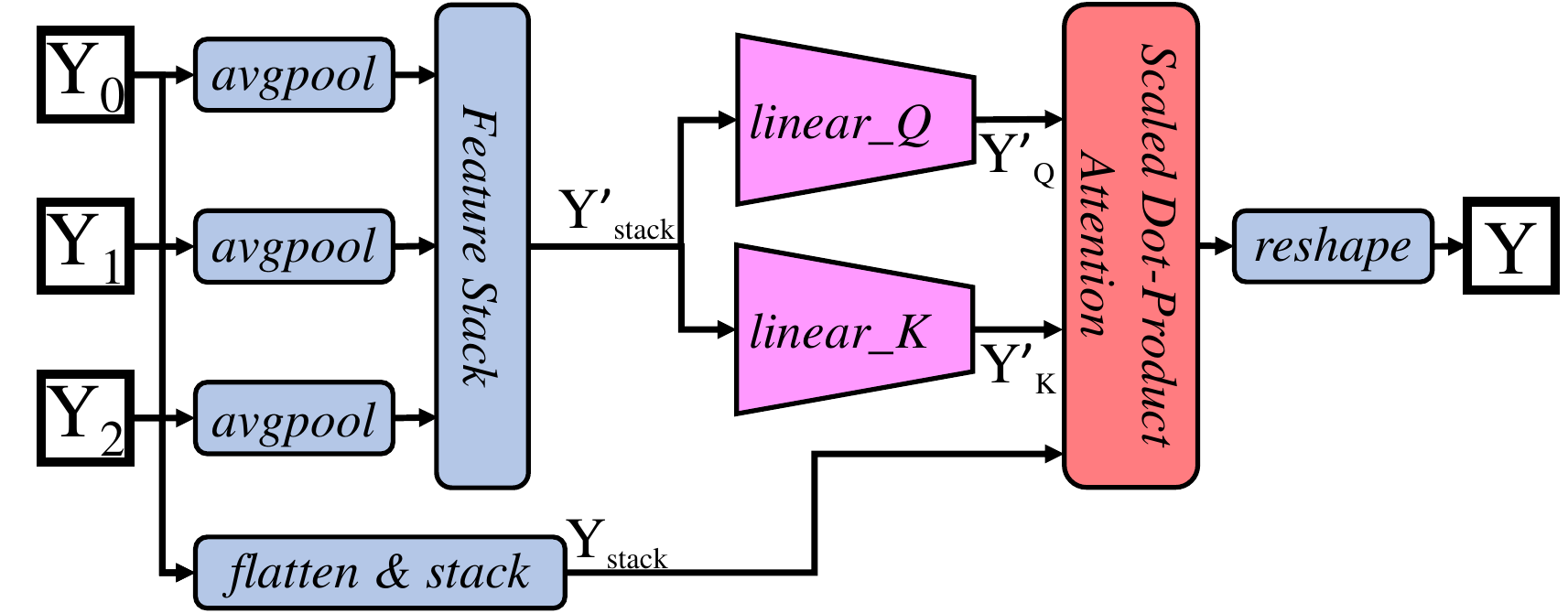}
    \caption{Structure of the combination function $f_{comb}$. $f_{comb}$ is designed by adopting a self-attention mechanism, and combines the multi-scaled dependencies attentively.}
    \label{figure:method:fcomb}
\end{figure}

\subsection{Pyramid Non-Local Module}
\label{method:pnl}
While the original non-local block is designed to capture long-range dependencies between any two positions in the input feature, such dependencies are extracted at the pixel level, where pixels at every space-time position is included in the computation. The use of only pixel correlation may not be effective and efficient due to the existence of trivial background pixels and the exclusion of regional correlation. On the other hand, multi-scale regional features have been proven effective in tasks such as object detection \cite{lin2017feature} and salient object detection \cite{wang2019salient}. Inspired by these works, we introduce the Pyramid Non-Local module (PNL) which incorporates multi-scale regional correlation, utilizing a pyramid structured module. The structure of the PNL module is as shown in Figure~\ref{figure:method:nl-pnl-b}. Formally, to extend the original non-local block to the regional level, we first obtain $n$ features of different scales from the original input $\mathbf{X}$. We leverage the channel grouping technique as in \cite{wu2018group,xie2017aggregated}, grouping the channels into $n$ groups, each containing $C^\prime=C/n$ channels, with $n$ strictly larger than 1. We denote the result of channel grouping to be $\mathbf{X}^\prime_0, \mathbf{X}^\prime_1$,..., $\mathbf{X}^\prime_{n-1}$. We then obtain the features of $n$ scales through sub-sampling operations over the $n$ groups of feature. Note that the sub-sampling operation does not apply to $\mathbf{X}^\prime_0$, where we preserve a group of channels with the same resolution and scale as the original input. The sub-sampling operations are implemented as pooling operation on the spatial dimensions. The result of the $k^{th}$ sub-sampling operation $\mathbf{X}_k$ is of spatial size $\frac{H}{2^k}\times \frac{W}{2^k}$. The results of the above channel grouping and sub-sampling process are therefore scaled features denoted as $\mathbf{X}_k\in\mathbb{R}^{T\times \frac{H}{2^k}\times \frac{W}{2^k}\times C^\prime}$, where $k\in[0, (n-1)]$. 

Through the pooling process, each feature point in the scaled features corresponds to a region of the original input. Therefore, the correlation of each feature point in the scaled features can be viewed as the correlation of the corresponding regions in the original input. To capture the long-range dependencies on both the pixel level and regional level, we input the scaled features of $\mathbf{X}_0, \mathbf{X}_1$,..., $\mathbf{X}_{n-1}$ to the non-local operation, as reviewed in Section~\ref{method:review-nl}. For all the input scaled features, we share the parameters of the non-local operations. The end result of this step are thus long-range dependencies at multiple scales, denoted as $\mathbf{Y}_0, \mathbf{Y}_1$,..., $\mathbf{Y}_{n-1}$. To obtain the overall long-range dependencies denoted as $\mathbf{Y}$, we combine the long-range dependencies of $\mathbf{Y}_0, \mathbf{Y}_1$,..., $\mathbf{Y}_{n-1}$ with a combination function $f_{comb}$. $f_{comb}$ could be simply a vanilla concatenate function, where $\mathbf{Y}=concat(\mathbf{Y}_0, \mathbf{Y}_1$,..., $\mathbf{Y}_{n-1})$. However, the vanilla concatenate function weighs all input features equally, which is not ideal. To combine the multi-scaled long-range dependencies dynamically, our proposed $f_{comb}$ adopt a self-attention mechanism, utilizing the scaled dot-product attention introduced in \cite{vaswani2017attention}. The structure of $f
_{comb}$ is presented in Figure~\ref{figure:method:fcomb}.

Given the multi-scaled long-range dependencies $\mathbf{Y}_0, \mathbf{Y}_1$,..., $\mathbf{Y}_{n-1}$, the scale-attended long-range dependencies $\mathbf{Y}$ is computed as:
\begin{equation}
\label{eqn:method:obtain-y}
\mathbf{Y} = Reshape(Attend(\mathbf{Y}^\prime_Q, \mathbf{Y}^\prime_K, \mathbf{Y}_{stack})),
\end{equation}
where the $Attend(\cdot)$ function is implemented as the scaled dot-product attention while $Reshape(\cdot)$ reshapes the output of the $Attention(\cdot)$ function to match that of the original input feature $\mathbf{X}\in\mathbb{R}^{T\times H\times W\times C}$. The $Attend(\cdot)$ function is formulated as:
\begin{equation}
\label{eqn:method:scaled-dot-attention}
Attend(\mathbf{Y}^\prime_Q, \mathbf{Y}^\prime_K, \mathbf{Y}_{stack}) = \sigma(\frac{\mathbf{Y}^\prime_Q{\mathbf{Y}^\prime_K}^T}{\sqrt{C^\prime}})\mathbf{Y}_{stack}.
\end{equation}
Here $\sigma(\cdot)$ is the softmax function, which ensures that the weights for all scales add up to 1. $\mathbf{Y}_{stack}$ is obtained through by first flattening long-range dependencies of all scales spatiotemporally and stacked along a separate "scale dimension". Both $\mathbf{Y}^\prime_Q$ and $\mathbf{Y}^\prime_K$ are obtained through a three step process: first, a spatiotemporal average pooling operation is applied to the multi-scaled long-range dependencies to obtain a representation for the dependencies of each scale; second, the pooled dependencies are stacked along the separate "scale dimension" to form a stacked representation feature, denoted as $\mathbf{Y}^\prime_{stack}$; third, separate trainable linear layers, $linear$\textunderscore $Q$ and $linear$\textunderscore $K$ are applied to $\mathbf{Y}^\prime_{stack}$ obtain $\mathbf{Y}^\prime_Q$ and $\mathbf{Y}^\prime_K$. The end product of Equation~\ref{eqn:method:scaled-dot-attention} and Equation~\ref{eqn:method:obtain-y} would be the overall long-range dependencies with dynamic weights applied to the dependencies of each scale.

\subsection{Computational Efficiency Analysis for PNL Module}
\label{method:proof-efficiency}
In this section, we prove the efficiency for extracting the long-range dependencies $\mathbf{Y}$ with our proposed PNL module against the original non-local block. In this proof, we adopt the case where $f(\cdot,\cdot)$ is the embedded Gaussian version. For notation simplicity, here we denote $N=T\times H\times W$. Under this notation, the dimension for input $\mathbf{X}\in\mathbb{R}^{T\times H\times W\times C}$ could be simplified as $\mathbf{X}\in\mathbb{R}^{N\times C}$. 

We first compute the computation cost for the original non-local block, though $conv$\textunderscore $\theta$, $conv$\textunderscore $\phi$ and $conv$\textunderscore $g$ operations are convolutional, they are essentially linear multiplicative operations. Their designed to project the original input to an embedding space with lower dimension. As designed in \cite{wang2018non}, the embedding space is of dimension $\mathbb{R}^{N \times \frac{C}{2}}$. Similarly, the operation $conv$\textunderscore $z$ as shown in Figure~\ref{figure:method:nl-pnl-a} is also a linear multiplicative which projects the computed dependencies back from the embedding space. The total computation cost for operations $conv$\textunderscore $\theta$, $conv$\textunderscore $\phi$, $conv$\textunderscore $g$ and $conv$\textunderscore $g$ could thus be computed as:
\begin{equation}
\label{eqn:method:cost-nl-conv1}
Cost_{nl,embs} = 4\times C^2\times N\times \frac{C}{2} = 2C^3N.
\end{equation}
The subsequent matrix multiplication of the embeddings from $conv$\textunderscore $\theta$ and $conv$\textunderscore $\phi$ would be computed as:
\begin{equation}
\label{eqn:method:cost-nl-mul1}
Cost_{nl,matmul} = \frac{C}{2}\times N\times N = \frac{1}{2}CN^2.
\end{equation}
The same computation cost also applies to the matrix multiplication between the softmax result of the previous matrix multiplication with the embeddings from $conv$\textunderscore $g$. The computation cost of softmax function is negligible compared to the multiplicative computations as listed above. The approximate total computation cost of the original non-local block is thus computed as:
\begin{equation}
\label{eqn:method:cost-nl-total}
\begin{split}
Cost_{nl} & = Cost_{nl,embs} + 2\times Cost_{nl,matmul}\\
& = 2C^3N + CN^2.
\end{split}
\end{equation}

We now consider the computation cost for our proposed PNL module, which utilizes the non-local operation while incorporating regional correlation. To compute the overall computation cost for PNL module, we first compute the computation cost for the process of obtaining $\mathbf{Y}_k$ from $\mathbf{X}_k$ as indicated in Figure~\ref{figure:method:nl-pnl-b}, denoted as $Cost_{nl, k}$. The computation of $Cost_{nl, k}$ 
follows the same procedures as that of the computation of $Cost_{nl}$. However they differ in two perspectives: first, the channel number of $\mathbf{X}_k$ is $C^\prime=C/n$ and second, as $\mathbf{X}_k$ is of dimension $\mathbb{R}^{T\times \frac{H}{2^k}\times \frac{W}{2^k}\times C^\prime}$, thus following the notation above, $N_k$ is computed as:
\begin{equation}
\label{eqn:method:cost-pnl-nk}
\begin{split}
N_k & = T\times \frac{H}{2^k}\times \frac{W}{2^k}\\
& = \frac{1}{2^{2k}}T\times H\times W \\
& = \frac{1}{4^{k}}N.
\end{split}
\end{equation}
Thus we could compute $Cost_{nl, k}$ as:
\begin{equation}
\label{eqn:method:cost-pnl-nlk}
\begin{split}
Cost_{nl, k} & = 2{C^\prime}^{3}N_K + {C^\prime}{N_k}^2\\
& = \frac{1}{n^3}2C^{3}\frac{1}{4^{k}}N + \frac{1}{n}C\times ({\frac{1}{4^{k}}N})^2 \\
& = \frac{1}{n^3\times4^{k}}2C^{3}N + \frac{1}{n\times 16^{k}}CN^2.
\end{split}
\end{equation}
Hence, the total computation cost of obtaining the multi-scaled long-range dependencies in our PNL module can be computed as:
\begin{equation}
\label{eqn:method:cost-pnl-dependencies}
\begin{split}
Cost_{pnl, dep} & = \sum\limits_{k=0}^{n-1}Cost_{nl, k}\\
& = (\sum\limits_{k=0}^{n-1}\frac{1}{4^{k}})\frac{1}{n^3}2C^{3}N + (\sum\limits_{k=0}^{n-1}\frac{1}{16^{k}})\frac{1}{n}CN^2.
\end{split}
\end{equation}
As the scale of the feature map must be a positive integer, it could be easily computed that $n$ could only take the values of 2, 3 or 4, where the largest $Cost_{pnl, dep}$ is obtained with $n=2$ with $Cost_{pnl, dep}=\frac{5}{32}2C^{3}N + \frac{17}{32}CN^2$. Meanwhile, the computation with regards to $f_{comb}$ is negligible compared to the computation cost of obtaining the dependencies. The above proof clearly proves that our proposed PNL is more efficient than the original non-local block in terms of lower computation cost.

\section{Experiments and Discussion}
\label{section:experiments}

In this section, we present the evaluation results of the proposed PNL module. The evaluation is conducted through action recognition experiments on two public benchmark datasets. We present state-of-the-art results on a competitive architecture. Further visualization results are also presented to justify the effectiveness of our proposed module.

\subsection{Experimental Settings}
\label{exp:settings}
\subsubsection{Datasets and Baselines}
For the action recognition task, we conduct experiments on two challenging public benchmark datasets: Mini-Kinetics \cite{xie2018rethinking} and UCF101 \cite{soomro2012ucf101}. The Mini-Kinetics is a subset of the Kinetics-400 \cite{kay2017kinetics} dataset, with 200 of its categories. It contains a total of 80K training data and 5K validation data. To obtain the state-of-the-art result on the Mini-Kinetics dataset, we instantiate MFNet \cite{chen2018multi} as the baseline thanks to its outstanding performance on Kinetics-400 dataset.

The UCF101 \cite{soomro2012ucf101} dataset contains 13,320 videos with 101 categories. For the UCF101 dataset, we follow the settings as in previous works \cite{chen2018multi,tran2018closer}, and adopt the three train/test splits for evaluation. We report the average top-1 accuracy over the three splits. Our proposed PNL module can be used with any current CNN networks. Due to the high performance of MFNet, the effectiveness of our proposed PNL module may not be obvious. Instead, thanks to its the steady performance, ResNet-50 \cite{he2016deep} is adopted as the baseline for experiments on the UCF101 dataset, denoted here as R-50. We adopt the exact same architecture configuration as in \cite{wang2018non}, where the temporal dimension is trivially addressed through pooling operation and the convolutional kernels are of size $1\times k\times k$.

\subsubsection{Implementation Details}
Our experiments are all implemented using PyTorch \cite{paszke2019pytorch}. Following the implementation in \cite{chen2018multi}, the input is a frame sequence with each frame of size $224\times224$. For the MFNet baseline, we follow the implementation in \cite{chen2018multi} and sample a sequence of 16 frames. Whereas for the ResNet-50 baseline, we sample a sequence of 32 frames as suggested in \cite{wang2018non}. To accelerate our training, we utilize the pretrained model of MFNet trained on Kinetics-400, and the pretrained model of ResNet-50 trained on ImageNet \cite{russakovsky2015imagenet}. The stochastic gradient descent algorithm \cite{bottou2010large} is used for optimization, with the weight decay set to 0.0001 and the momentum set to 0.9. Our initial learning rate is set to 0.01. Similar to \cite{wang2018non}, we ensure that the initial state of the entire PNL module to be an identity mapping. This further ensures that the proposed PNL modules can be inserted into any pretrained networks while maintaining its initial behavior.

\subsection{Ablation Experiments}
\label{experiment:ablation}
We obtain an optimal form of PNL while verifying our design through ablation experiments. The ablation experiments are all conducted on the Mini-Kinetics dataset utilizing the MFNet baseline.

\begin{table}
\centering
\resizebox{.5\textwidth}{!}{
\smallskip\begin{tabular}{c|c|c}
    \hline
    \hline
    Pairwise Function & Top-1 & Top-5\\
    \hline
    MFNet baseline & 78.35\% & 94.65\%\\
    \hline
    Embedded Gaussian & 82.16\% & 95.83\%\\
    Gaussian & 81.68\% & 95.51\%\\
    Dot Product & 81.45\% & 95.54\%\\
    Concatenation & 81.79\% & 95.36\%\\
    \hline
    \hline
\end{tabular}
}
\caption{\textbf{Ablation 1 - Type of pairwise function:} A single PNL module with $n=4$ with different types of pairwise function $f(\cdot,\cdot)$ is inserted into the MFNet baseline. All are inserted to the last multi-fiber unit right before the end of $conv4$ stage.}\label{table:exp:pairwise} 
\end{table}
\begin{table}
\centering
\resizebox{.5\textwidth}{!}{
\smallskip\begin{tabular}{c|c|c}
    \hline
    \hline
    PNL position & Top-1 & Top-5\\
    \hline
    MFNet baseline & 78.35\% & 94.65\%\\
    \hline
    $conv2$ & 81.41\% & 95.33\%\\
    $conv3$ & 81.63\% & 95.48\%\\
    $conv4$ & 81.98\% & 95.59\%\\
    $conv5$ & 81.37\% & 95.31\%\\
    \hline
    \hline
\end{tabular}
}
\caption{\textbf{Ablation 2 - Position of PNL:} A single PNL module with $n=2$ is inserted into the MFNet baseline. The insertion is located at the last multi-fiber unit right before the end of each stage.}\label{table:exp:position} 
\end{table}
\begin{table}
\centering
\resizebox{.8\textwidth}{!}{
\smallskip\begin{tabular}{c|c|c|c|c}
    \hline
    \hline
    Combination Function & Top-1 & Top-5 & \# Params & Flops\\
    \hline
    MFNet baseline & 78.35\% & 94.65\% & 7.843M & 11.176G\\
    \hline
    Vanilla concatenation & 81.93\% & 95.37\% & 7.911M & 11.208G\\
    Self-attention mechanism & 82.16\% & 95.83\% & 7.92M & 11.218G\\
    \hline
    \hline
\end{tabular}
}
\caption{\textbf{Ablation 3 - Type of combination function:} A single PNL module with $n=4$ is inserted into the MFNet baseline at the last multi-fiber unit right before the end of $conv4$. The multi-scaled long-range dependencies are combined with different types of $f_{comb}$.}\label{table:exp:fcomb} 
\end{table}
\begin{table}
\centering
\resizebox{.5\textwidth}{!}{
\smallskip\begin{tabular}{c|c|c}
    \hline
    \hline
    $n$ scales & Top-1 & Top-5\\
    \hline
    MFNet baseline & 78.35\% & 94.65\%\\
    \hline
    2 & 81.98\% & 95.59\%\\
    3 & 82.14\% & 95.74\%\\
    4 & 82.16\% & 95.83\%\\
    \hline
    \hline
\end{tabular}
}
\caption{\textbf{Ablation 4 - Number of scales:} A single PNL module with different scales of dependencies is inserted into the MFNet baseline at the same position.}\label{table:exp:scales} \smallskip
\end{table}

\subsubsection{Pairwise Function}
We first discuss the effect of the pairwise function $f(\cdot,\cdot)$ in the non-local block. Following \cite{wang2018non}, we utilize four types of pairwise functions, namely embedded Gaussian, Gaussian, dot product and concatenation. The result of utilizing each pairwise function is as shown in Table~\ref{table:exp:pairwise}. Consistent improvements can be observed regardless of the pairwise function utilized. Among which, the embedded Gaussian function as depicted in Figure~\ref{figure:method:nl-pnl-a} achieves the best performance. Therefore, the pairwise function $f(\cdot,\cdot)$ would be the embedded Gaussian version by default for the rest of the experiments.

\subsubsection{Position of PNL Module}
Table~\ref{table:exp:position} compares the result where a single PNL module is inserted to the different stages of the MFNet baseline. Note that due to the constraint imposed by the size of the feature map, the inserted PNL module includes only $n=2$ scales. The improvement of adding the PNL module gradually increases with the PNL module inserted into deeper stages until $conv4$ stage. However, the improvement by adding PNL module decreases sharply when then PNL module is inserted at the $conv5$ stage. The fact that the spatial dimension of feature map at $conv5$ stage is too small ($7\times7$) such that precise spatial dependencies could not be obtained even at the original feature map scale could be a reason of the sudden drop in improvement. This phenomena is inline with that observed in \cite{wang2018non}, where inserting the original non-local block at the last convolution stage also results in the lowest improvement. Thanks to the best performance obtained by inserting at the $conv4$ stage, we insert PNL module right before the last multi-fiber unit of $conv4$ stage by default. The multi-fiber unit in the MFNet baseline is equivalent to a residual block in the ResNet baseline.

\subsubsection{Combining Multi-scaled Dependencies with $f_{comb}$}
As mentioned in Section~\ref{method:pnl}, the multi-scaled dependencies obtained from the multi-scaled features are combined with a combination function $f_{comb}$. Here we compare the result utilizing two different types of combination function: a vanilla concatenation function, and a function utilizing self-attention mechanism as proposed in Section~\ref{method:pnl}. The results are presented in Table~\ref{table:exp:fcomb}. In addition to the Top-1 and Top-5 accuracies, we also compare the number of parameters and required computation Flops with respect to the different combination functions. It can be seen that our proposed self-attended $f_{comb}$ outperforms the vanilla concatenation combination by $0.26\%$. This is at a cost of only $0.09M$ extra parameters, which is less than $0.12\%$ increase in parameter size. This indicates that our proposed $f_{comb}$ is both effective and efficient, with a negligible computation cost.

\subsubsection{Number of Scales}
Table~\ref{table:exp:scales} shows the result of utilizing different numbers of scales in the PNL module. Due to the limitations of the scale of feature map, the number of scales is limited to a maximum number of $n=4$. Note that when $n=1$, the PNL module would be exactly same as the original non-local block. Hence we would not discuss the case where $n=1$. The results in Table~\ref{table:exp:scales} shows that with the increase in number of scales, the improvement would slightly increase. From Section~\ref{method:proof-efficiency}, it is also clear that with the increase in $n$, the computation cost of PNL module decreases. Hence when $n=4$ scales are utilized, our proposed PNL module is both effective and efficient. For the rest of the experiments, $n$ would be set to 4 by default.

\begin{table*}[t]
    \centering
    \small
    \setlength{\tabcolsep}{4pt}
    \resizebox{1.\textwidth}{!}{
    \smallskip\begin{tabular}{c|c|c|c|c}
        \hline
        \hline
        & Method & Mini-Kinetics Top-1 & \# Params & FLOPs\\
        \hline
        \multirow{3}{*}{\parbox{2cm}{\centering Two-stream CNNs}}
        & MARS \cite{crasto2019mars}                & 73.5\% & - & -\\
        & ResFrame TS \cite{tao2020rethinking}      & 73.9\% & - & -\\
        & I3D (TS) \cite{carreira2017quo}           & 78.7\% & 25.0M & $>\!107.9$G\\
        \hline
        \multirow{5}{*}{3D CNNs}
        & C3D \cite{tran2015learning}               & 66.2\% & 33.3M & -\\
        & I3D (RGB) \cite{carreira2017quo}          & 74.1\% & 12.06M & 107.9G\\
        & (2+C1)D \cite{cheng2019sparse}            & 75.74\% & \textbf{7.3M} & 31.9G\\
        & S3D \cite{xie2018rethinking}              & 78.0\% & 8.77M & 43.47G\\
        & MFNet \cite{chen2018multi}                & 78.35\% & 7.84M & \textbf{11.17G}\\
        \hline
        \multirow{4}{*}{\parbox{3cm}{\centering CNN with long-range dependencies}}
        & Res50-NL \cite{wang2018non}               & 77.53\% & 27.66M & 19.67G\\
        & Res50-CGD \cite{he2019compact}            & 77.56\% & 25.58M & 17.88G\\
        & Res50-CGNL \cite{yue2018compact}          & 77.76\% & 27.2M & 19.16G\\
        & MFNet-NL \cite{yue2018compact}            & 79.74\% & 8.15M & 11.66G\\
        \hline
        \multirow{2}{*}{Ours}
        & \textbf{MFNet-PNL($\times1$)}             & 82.16\% & 7.92M & 11.22G\\
        & \textbf{MFNet-PNL($\times5$)}             & \textbf{83.09\%} & 8.12M & 11.38G\\
        \hline
        \hline
    \end{tabular}
    }
    \caption{Comparison of top-1 and top-5 accuracy, number of parameters and computation cost in FLOPs with state-of-the-art methods on the Mini-Kinetics datasets.}\smallskip
    \label{table:exp:mk200}
\end{table*}

\begin{table}
\centering
    \resizebox{.8\columnwidth}{!}{
    \smallskip\begin{tabular}{c|c|c|c|c}
    \hline
    \hline
    Method & Top-1 & Top-5 & \# Params & Flops\\
    \hline
    R-50 & 81.62\% & 94.62\% & 23.92M & 10.29G\\
    \hline
    R-50 + NL & 82.88\% & 95.74\% & 26.38M & 18.72G\\
    R-50 + CGNL & 83.38\% & 95.42\% & 26.22M & 18.23G\\
    R-50 + PNL($\times1$) & 85.22\% & 95.82\% & 24.46M & 13.31G\\
    \hline
    \hline
    \end{tabular}
    }
\caption{Comparison of top-1 and top-5 accuracy, number of parameters and computation cost in FLOPs of ResNet-50 and its variants on the UCF101 dataset. The parameter size and computation FLOPs are lower for the same network than that tested on Mini-Kinetics due to the fewer number of classes.}
\label{table:exp:ucf101}
\end{table}

\subsection{Results and Comparison}
\label{exp:results}

Table~\ref{table:exp:mk200} shows the comparison of top-1 accuracy on Mini-Kinetics dataset with other current state-of-the-art methods which includes the following methods:
\begin{enumerate}
        \item \textit{Two-stream CNN methods: }MARS \cite{crasto2019mars}, Residual Frame with two-stream input (ResFrame TS) \cite{tao2020rethinking} and I3D with two-stream input \cite{carreira2017quo}.
        \item \textit{3D CNN methods: }C3D \cite{tran2015learning}, I3D with RGB input \cite{carreira2017quo}, (2+C1)D \cite{cheng2019sparse}, MFNet \cite{chen2018multi} and S3D \cite{xie2018rethinking}. 
        \item \textit{CNN with long-range dependencies: }Res50-NL \cite{wang2018non}, Res50-CGD \cite{he2019compact}, Res50-CGNL \cite{yue2018compact} and MFNet with non-local block inserted (MFNet-NL).
\end{enumerate}
The above methods are compared with MFNet-PNL($\times1$) which includes only a single PNL module with the MFNet baseline, and MFNet-PNL($\times5$) which includes five PNL modules. For the single PNL module case, the PNL module is inserted right before the last multi-fiber unit of $conv4$ stage of the MFNet baseline. For the five PNL modules case, the PNL modules are inserted to every other multi-fiber unit of $conv3$ and $conv4$ stage of the MFNet baseline. For this experiment, we set our batch size to 64 for the Mini-Kinetics dataset and conduct the experiment using two NVIDIA GP100 GPUs.

The results in Table~\ref{table:exp:mk200} clearly show that with the addition of our proposed PNL module, the network achieves the best result on the Mini-Kinetics dataset with limited increase in the number of parameters and computation cost compared to the original MFNet baseline. By inserting a single PNL module, the network achieves a $3.81\%$ increase over the baseline model. Utilizing the PNL module also outperforms the network with the same MFNet baseline but utilizing the original non-local block, denoted as MFNet-NL. In contrast, a single PNL module has 0.22M less parameters and requires 0.42G less FLOPs compared to the original non-local block. The optimal network performance on Mini-Kinetics is obtained by adding five PNL modules, increasing the accuracy by $3.35\%$ compared to the baseline. It can be noted that even with five PNL modules added, the total number of parameters and required computation FLOPs are both lower than that with the original non-local block. This further proves the effectiveness and efficiency of our proposed PNL module.

\begin{figure*}[!t]
    \centering
    \includegraphics[width=1.\linewidth]{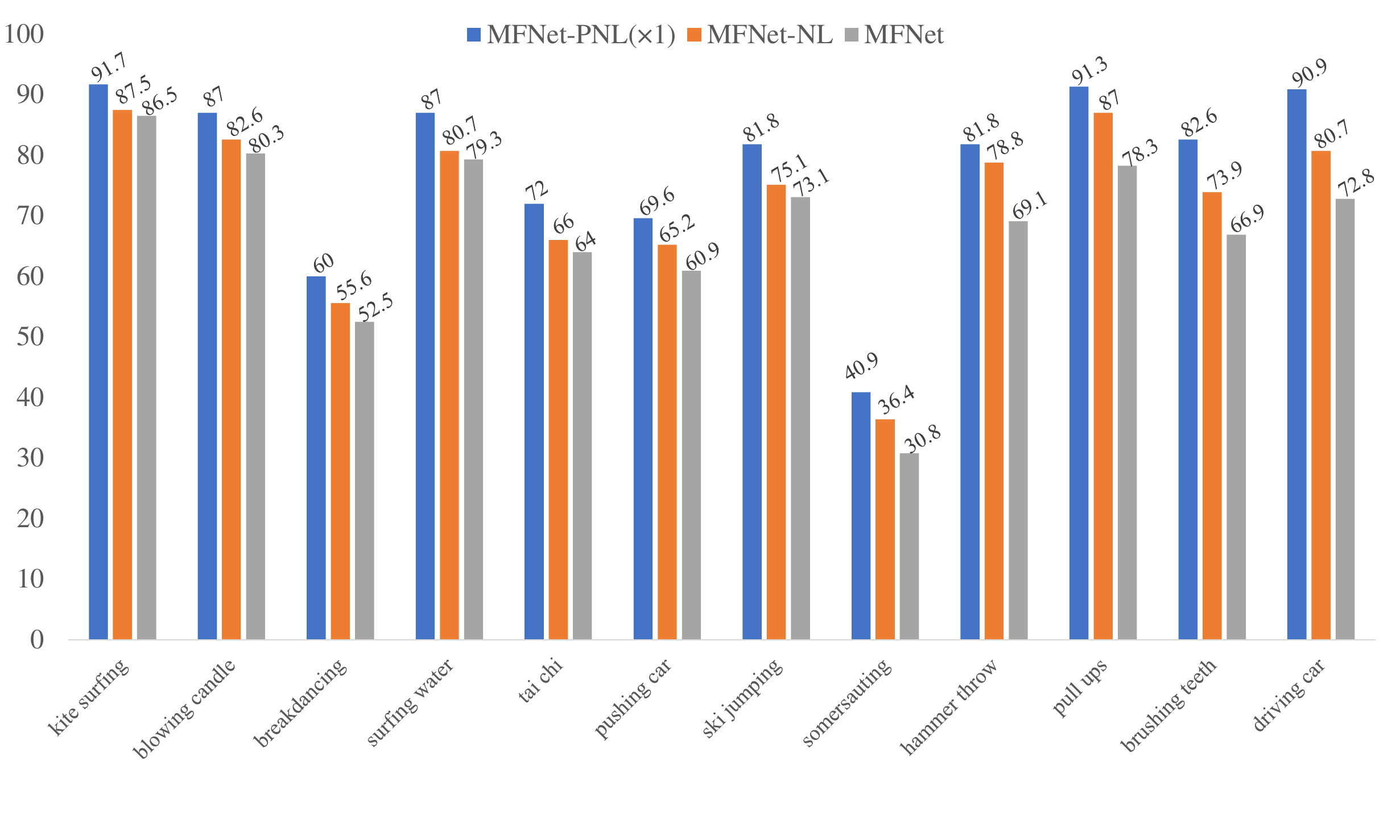}
    \caption{Detailed comparison of accuracy per class on the Mini-Kinetics between the original MFNet baseline with networks resulting from insertion of a single PNL module (MFNet-PNL($\times1$)) or a single non-local block (MFNet-NL). Here we present the accuracies of 12 classes where MFNet-PNL($\times1$) outperforms the original MFNet baseline by a margin of at least $5\%$. In all classes presented, the MFNet-PNL($\times1$) also outperforms MFNet-NL.}
    \label{figure:exp:perclass}
\end{figure*}

\begin{figure*}[!t]
    \centering
    \includegraphics[width=.9\linewidth]{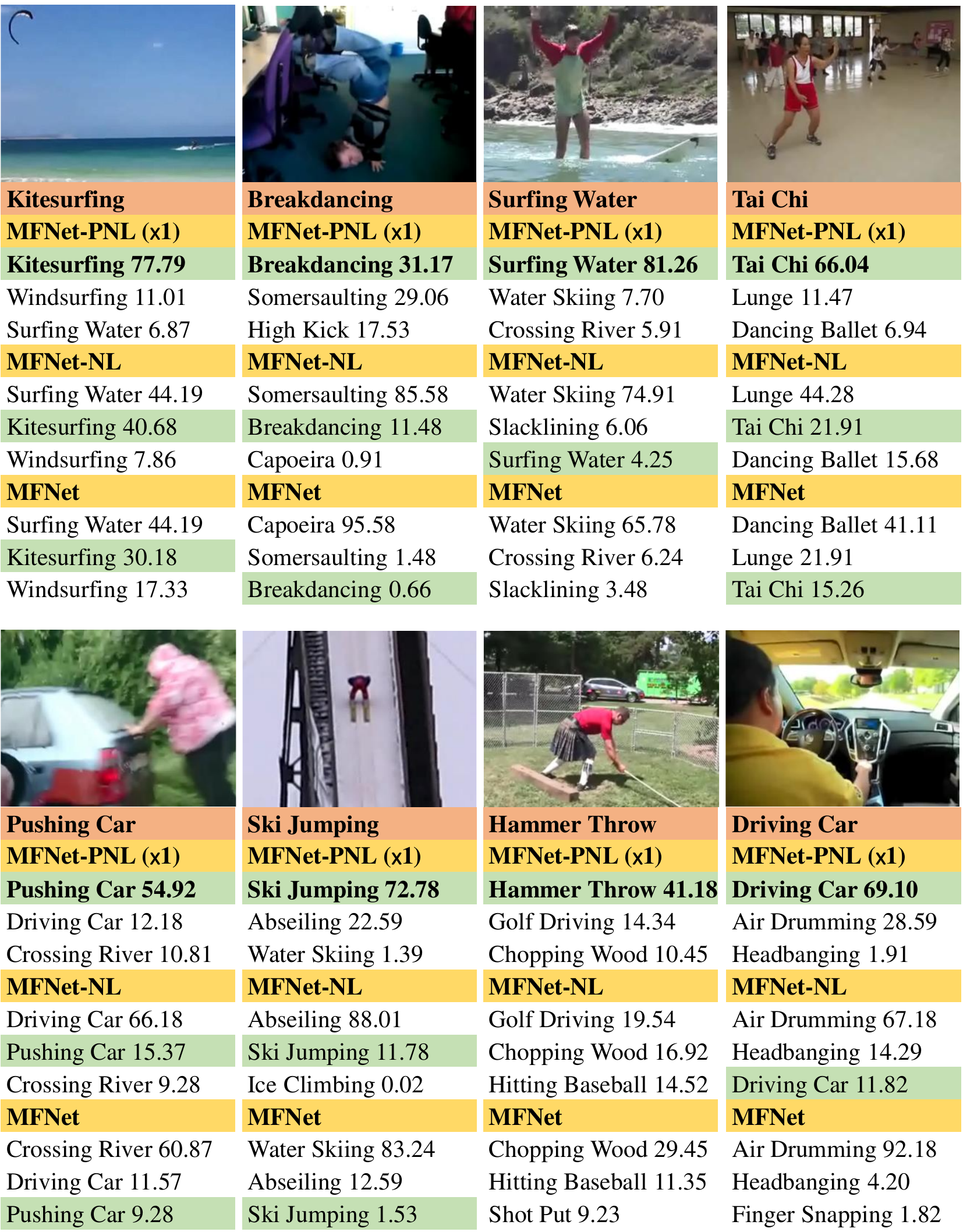}
    \caption{Eight examples taken from the 12 classes presented in Figure~\ref{figure:exp:perclass}. The numbers on the right of each class shows the probability of the class from the classifier in percentages. We show three classes with highest probability. The class with the highest probability is the result of the top-1 classification, highlighted in green.}
    \label{figure:exp:examples}
\end{figure*}

Besides testing on the Mini-Kinetics dataset, we also conduct experiments on the UCF101 dataset. Here we utilize the simpler ResNet-50 baseline instead of the MFNet baseline to showcase the effectiveness of the PNL module. The result is as presented in Table~\ref{table:exp:ucf101}. Here a single non-local block or PNL module is inserted at the exact same location, which is right before the last residual block of $res4$ stage. By comparison, inserting the proposed PNL module brings an extra $2.34\%$ increase in top-1 accuracy. At the same time, our PNL module has 1.92M less parameters and requires 5.41G less FLOPs compare to the original non-local block. The above results further justifies the effectiveness and efficiency of our PNL module compared to the original non-local block.

We further investigate the improvement over different actions and present the comparison of performance between the baseline MFNet network with that of inserted a single non-local block or a single PNL module. Figure~\ref{figure:exp:perclass} shows the accuracy of 12 classes from the Mini-Kinetics dataset, where inserting our proposed PNL module outperforms the original baseline network by a noticeable margin of over $5\%$. Inserting the PNL module also outperforms that of inserting the non-local block in all of the 12 classes presented. To further illustrate the effectiveness of our PNL module, we present several examples in Figure~\ref{figure:exp:examples} where inserting a single PNL module to the original baseline outperforms the baseline network with or without non-local block inserted. The superior performance over inserting the non-local block in these examples illustrates that modeling regional correlation in long-range dependencies could bring additional information to the network, thus resulting in more effective video features.

\begin{figure*}[!t]
    \centering
    \subfloat[Visualizing Action ``Jetskiing"\label{figure:exp:visualize-1}]{
    \includegraphics[width=.9\textwidth]{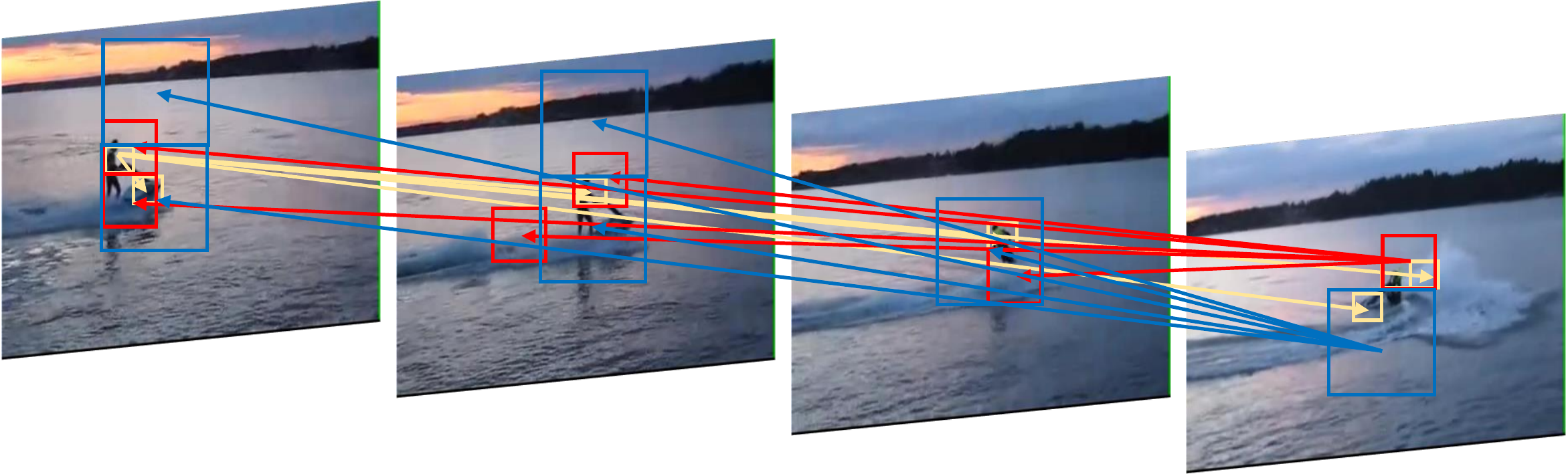}}\\\smallskip
    \centering
    \subfloat[Visualizing Action ``Kitesurfing"\label{figure:exp:visualize-2}]{
    \includegraphics[width=.9\textwidth]{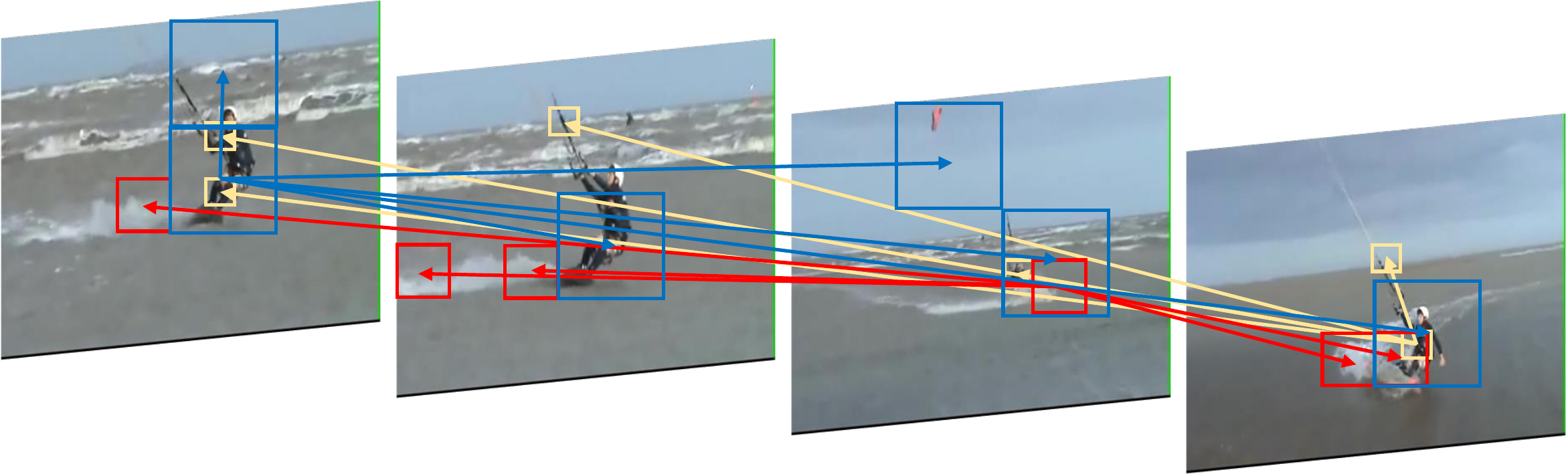}}\\\smallskip
    \centering
    \subfloat[Visualizing Action ``Passing Football"\label{figure:exp:visualize-3}]{
    \includegraphics[width=.9\textwidth]{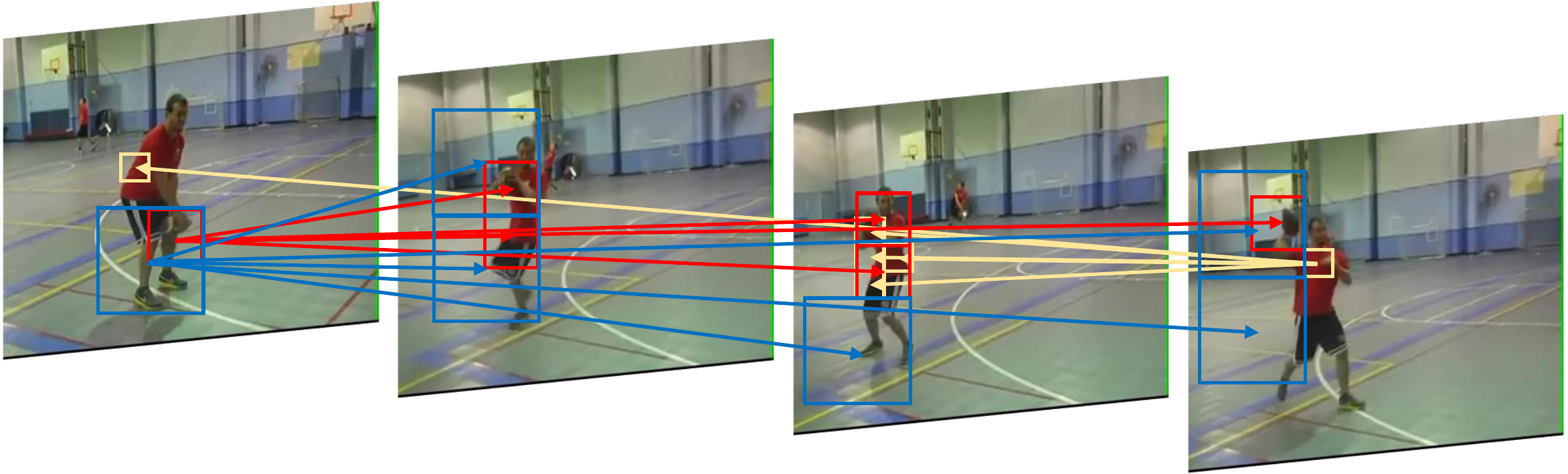}}\\\smallskip
    \centering
    \subfloat[Visualizing Action ``Rock Climbing"\label{figure:exp:visualize-4}]{
    \includegraphics[width=.9\textwidth]{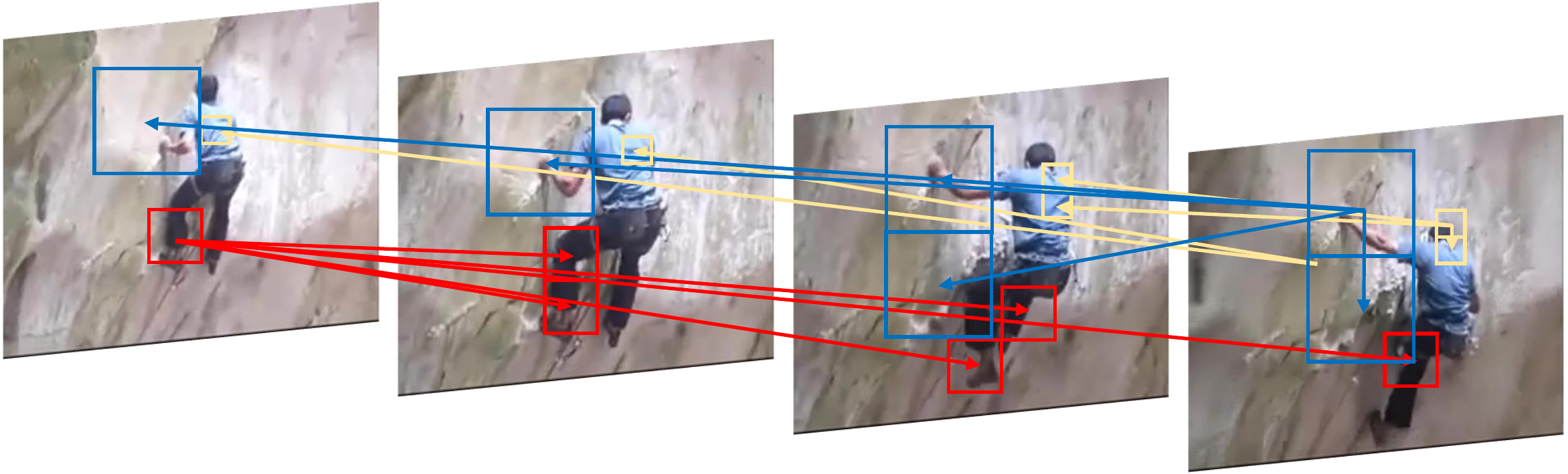}}\\
    \caption{Visualization of the behaviour of our PNL module. From a reference region, we visualize the five of the most correlated regions computed from PNL module at three different scales, shown in different colors. We observe that these correlations complements each other, thus capturing more effective long-range spatiotemporal dependencies. Figure best viewed in color and zoomed in.}
    \label{figure:exp:visualize}
\end{figure*}

\subsection{Visualization}
\label{exp:visualization}

To justify the effectiveness of our proposed module in capturing regional long-range dependencies at multiple scales, we visualize the interactions of the different regions in sample videos. Here for simplicity, we utilize the MFNet-PNL($\times1$) network. The visualization of the behaviour of our PNL module is as shown in Figure~\ref{figure:exp:visualize}. It could be observed that the multi-scaled long-range dependencies complements each other, providing effective information towards the final classification. For example, for the action ``Kitesurfing" in Figure~\ref{figure:exp:visualize-2}, the smallest scale long-range dependencies, obtained through the original feature map, captures the correlation between the person, the board underneath and the rope above. Whereas the largest scale long-range dependencies, obtained through the sub-sampled feature map, captures the correlation between the person and the kite above. Without this correlation, the action may be mis-classified with similar actions such as ``windsurfing", which is presented in a similar video in Figure~\ref{figure:exp:examples}.

\section{Conclusion and Future Works}
\label{section:conclusion}

In this work, we propose a novel module for effective capturing of long-range spatiotemporal dependencies. The proposed PNL module extends the original non-local block by incorporating regional correlation at multiple scales through a pyramid structural design. Our method obtains state-of-the-art result on the Mini-Kinetics dataset when instantiating MFNet, while bringing significantly less computation cost than the original non-local block. We further justify the design of our PNL module through detailed ablation study. We further demonstrate the effectiveness of the PNL module by visualizing the captured dependencies in sampled videos. 

In the future, the application of the PNL module to other video-based tasks, such as object tracking or video description could be explored. Capturing long-range feature dependencies plays an essential role in these tasks. Given the effectiveness of our PNL module in addressing such a challenge, we believe that applying PNL module could improve the performance of networks in various video-based tasks.

\clearpage
\bibliography{neuro}

\end{document}